\documentclass[letterpaper, 10 pt, conference]{ieeeconf}  

\IEEEoverridecommandlockouts                              

\usepackage{cite}

\usepackage{authblk}
\usepackage{graphics} 
\usepackage{epsfig} 
\usepackage{mathptmx} 
\usepackage{times} 
\usepackage{amsmath} 
\usepackage{amssymb}  
\usepackage{cite}
\usepackage{xcolor}
\usepackage{booktabs}
\usepackage{float}
\usepackage{multirow}
\usepackage{caption}
\usepackage{newtxtext, newtxmath}
\usepackage{tabularx}
\usepackage{longtable}

\usepackage{flushend}
\usepackage{amsmath}
\usepackage{stfloats}
\usepackage{balance}
\usepackage{xspace}
\usepackage{listings}
\usepackage{xcolor}  
\usepackage{algorithm}
\usepackage{algpseudocode}
\usepackage{tcolorbox}
\usepackage[pagebackref=true,breaklinks=true,colorlinks,citecolor=gray]{hyperref}

\title{\LARGE \bf
MoE-Loco: Mixture of Experts for Multitask Locomotion
}

\author{Runhan Huang*$^{1,2}$, Shaoting Zhu*$^{1,2}$, Yilun Du$^{3}$, Hang Zhao†$^{1,2}$\\
\href{https://moe-loco.github.io/}{\large\textbf{MoE-Loco.github.io/}}
\thanks{$^{1}$IIIS, Tsinghua University, Beijing, China}%
\thanks{$^{2}$Shanghai Qi Zhi Institute, Shanghai, China}%
\thanks{$^{3}$Massachusetts Institute of Technology, MA, USA}%
\thanks{* These authors contributed equally to this work.}
\thanks{† Corresponding at: \texttt{hangzhao@mail.tsinghua.edu.cn}}%
}

\begin{document}

\newcommand{\insertteaser}{
    \includegraphics[width=0.85\linewidth]{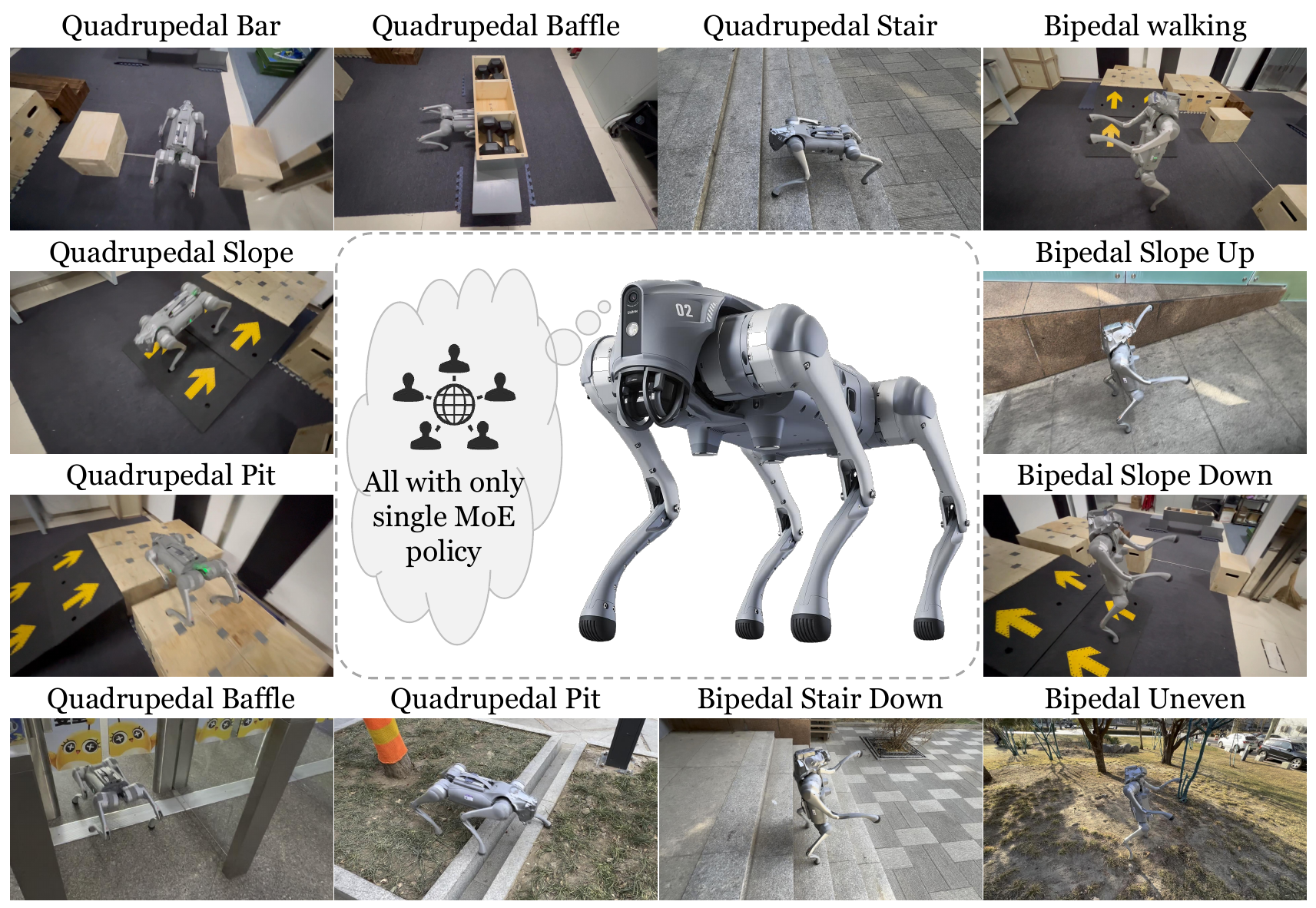}
    \vspace{-1mm}
    \captionof{figure}{\textbf{We introduce \texttt{MoE-Loco}.} With a single MoE policy, quadruped robot can traverse a variety of challenging terrains and perform different locomotion modes, including bipedal and quadrupedal gaits.}
    \vspace{-2mm}
    \label{fig: teaser}
}

\makeatletter
\apptocmd{\@maketitle}{\centering\insertteaser}{}{}
\makeatother

\maketitle
\setcounter{figure}{1}

\thispagestyle{empty}
\pagestyle{empty}

\begin{abstract}

We present \texttt{MoE-Loco}, a Mixture of Experts (MoE) framework for multitask locomotion for legged robots. Our method enables a single policy to handle diverse terrains, including bars, pits, stairs, slopes, and baffles, while supporting quadrupedal and bipedal gaits. Using MoE, we mitigate the gradient conflicts that typically arise in multitask reinforcement learning, improving both training efficiency and performance. Our experiments demonstrate that different experts naturally specialize in distinct locomotion behaviors, which can be leveraged for task migration and skill composition. We further validate our approach in both simulation and real-world deployment, showcasing its robustness and adaptability.

\end{abstract}

\section{INTRODUCTION}

Robots are often required to traverse diverse terrains and demonstrate various skills \cite{kumar2023cascaded, klipfel2023learning}. Recent advancements in reinforcement learning (RL) algorithms and physics-based simulators have enabled RL-based approaches to become the dominant paradigm for training robot locomotion policies \cite{kumar2021rma, su2024leveraging, ji2022concurrent, long2024learning, zhuang2023robot}. 
However, while single-task RL has demonstrated remarkable success, learning a unified policy that generalizes across multiple tasks, terrains, and locomotion modes remains a significant challenge.

Recent research has explored training locomotion policies with diverse skills by having diverse terrains in simulation for parallel training \cite{wu2023learning, luo2024pie, cheng2024extreme}. However, multitask RL with a simple neural network architecture often suffers from gradient conflicts \cite{liu2023improving, zhou2022convergence}, which leads to inferior model performance. What is worse, training a policy across multiple terrains with different gaits poses further challenges, leading to model divergence.

In this work, we enable a quadruped robot to traverse various terrains—including bars, pits, stairs, slopes, and baffles—while also supporting gait switching between bipedal and quadrupedal modes, using only one policy. We integrate the Mixture of Experts (MoE) framework \cite{jacobs1991adaptive, obando2024mixtures, li2024mixtures, celik2024acquiring} as a modular network structure for multitask locomotion reinforcement learning. We demonstrate that the MoE framework alleviates gradient conflicts by directing gradients to specialized experts, thus improving training efficiency and overall performance. Furthermore, we analyze the roles of different experts within the MoE model and observe that they naturally specialize in distinct behaviors. Leveraging this property, we can manually adjust the ratios between different experts to compose new skills, underscoring the adaptability and reusability of our modular approach for novel tasks.

In summary, our contributions are as follows.
\begin{itemize}
    \item We train and deploy a single neural network policy that enables a quadruped robot to \textbf{cross challenging terrains} and perform fundamentally different locomotion modes, including \textbf{bipedal and quadrupedal gaits}.
    \item We integrate the \textbf{MoE architecture} into locomotion policy training to \textbf{mitigate gradient conflicts}, improve training efficiency, and overall model performance.
    \item We conduct qualitative and quantitative analysis of MoE, uncovering expert specialization patterns. Using these insights, we explore the potential of MoE for \textbf{task migration} and \textbf{skill composition}.
\end{itemize}
\section{RELATED WORKS}

\subsection{Reinforcement Learning for Robot Locomotion}
Using reinforcement learning for robot locomotion control has become increasingly popular in recent years. It has demonstrated the ability to learn legged locomotion behaviors in both simulation~\cite{makoviychuk2021isaac, mittal2023orbit} and the real world~\cite{kumar2021rma, hwangbo2019learning}. Not only can it traverse a variety of complex terrains~\cite{wu2023learning, zhu2024robust, lee2020learning}, but it can also achieve high-speed running~\cite{margolis2024rapid, he2024agile}. Furthermore, reinforcement learning has enabled robots to perform extreme tasks and master skills such as bipedal walking~\cite{li2024learning, smith2023learning}, opening doors~\cite{su2024leveraging, kumar2023cascaded}, navigating rocky terrains~\cite{cheng2024quadruped}, and even executing high-speed parkour in challenging environments~\cite{zhuang2023robot, cheng2024extreme, hoeller2024anymal, luo2024pie}. However, most of these works focus on specific skills and a limited number of terrains, rarely considering multitask learning.

\subsection{MultiTask Learning}
Multitask learning (MTL) aims to train a unified network that can perform across different tasks~\cite{caruana1997multitask, collobert2008unified, vandenhende2021multi}. MTL allows multiple tasks to benefit from shared knowledge~\cite{liu2016recurrent, pinto2017learning}, but some works highlight the challenge of negative gradient conflicts during training~\cite{yu2020gradient, liu2021conflict, huang2024mentor}. Multitask Reinforcement Learning (MTRL) is one of the most popular areas of research in this domain. Many algorithms have been developed to improve the effectiveness of MTRL~\cite{finn2017model, duan2016rl, sodhani2021multi}. Moreover, MTRL is widely applied in the robotics field, though much of the focus has been on manipulation tasks~\cite{celik2024acquiring, huang2024mentor, yang2020multi, ze2023gnfactor}. In the context of locomotion, works like ManyQuadrupeds~\cite{shafiee2024manyquadrupeds} focus on learning a unified policy for different categories of quadruped robots. MELA~\cite{yang2020multiloco} employs pretrained expert models to construct a locomotion policy, although it primarily concentrates on basic skill acquisition. Moreover, the pretraining process for specialized neural network policies requires substantial reward engineering efforts. MTAC~\cite{shah2023mtac} attempts to train a policy across different terrains using hierarchical RL, but their approach can only handle one gait with three relatively simple terrains and has not been deployed on a real robot.

\subsection{Mixture of Experts (MoE)}
The concept of Mixture of Experts (MoE), originally introduced in \cite{jacobs1991adaptive, jordan1994hierarchical}, has received extensive attention in recent years \cite{deisenroth2015distributed, lepikhin2020gshard}. It has found widespread application in fields such as natural language processing \cite{jiang2024mixtral, liu2024deepseek}, computer vision \cite{zhang2024m3oe, jiang2023adamct}, and multi-modal learning \cite{gou2023mixture, chen2024llava}. MoE has also been applied in reinforcement learning and robotics. DeepMind \cite{obando2024mixtures} has explored using MoE to scale reinforcement learning. MELA \cite{yang2020multi} proposed a Multi-Expert Learning Architecture to generate adaptive skills from a set of representative expert skills, but their focus is primarily on simple actions.

\section{METHOD}

\subsection{Task Definition}

\begin{figure}[htbp]
    \centering
    \vspace{-2mm}
    \includegraphics[width=0.47\textwidth]{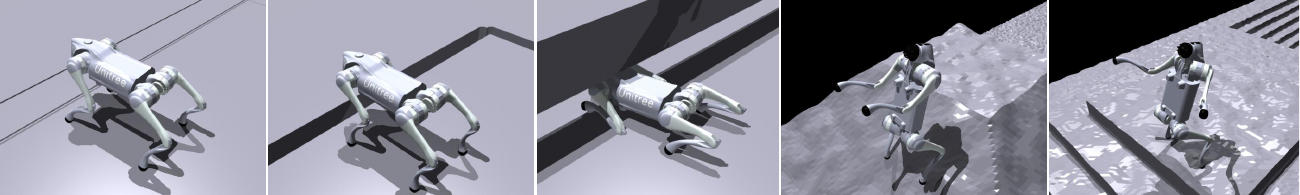}
    \caption{\textbf{A snapshot of the terrain settings.} From left to right: bar, pit, baffle, slope, and stairs.}
    \label{fig:task_def}
    \vspace{-3mm}
\end{figure}

In this paper, we focus on 9 challenging locomotion tasks, encompassing both quadrupedal and bipedal gaits. The quadrupedal gait tasks include bar crossing, pit crossing, baffle crawling, stair climbing, and slope walking. The bipedal gait tasks consist of standing up, plane walking, slope walking, and stair descending. Our terrains in the simulation environment are shown in \autoref{fig:task_def}. The robot is controlled via velocity commands from a joystick, where a one-hot vector is used to indicate whether to walk in the quadrupedal or bipedal gait.  

We define multitask locomotion as a Markov Decision Process (MDP), represented by the tuple $\langle S_{\tau}, A_{\tau}, T_{\tau}, R_{\tau}, \gamma_{\tau} \rangle$. 
In our training framework, the multitask nature is characterized by several key aspects. First, different locomotion terrains correspond to distinct subsets of the state space, denoted as $S_\tau$, where $S_\tau \subseteq S$ represents the states relevant to a specific task. However, the full state space $S$ remains unknown to the robot. For instance, the task of walking on slopes involves a state space that differs from those required for stair climbing or bar traversal.

Moreover, the reward function $R$ varies across different gaits, reflecting task-specific objectives. Additionally, the termination conditions depend on the type of gait, leading to distinct transition dynamics $T$. The robot learns a policy $\pi(a|s)$ that selects actions based on both the terrain and gait, aiming to maximize the cumulative reward across tasks:

\begin{equation}
J(\pi) = \mathbb{E} \left[ \sum_{\tau}\sum_{t=0}^{\infty} \gamma^{t} R_{\tau}(s_t, a_t, s_{t+1}) \right]
\end{equation}

The goal is to learn a single universal policy that generalizes across various tasks. We demonstrate our method in the condition of blind locomotion (only use proprioception as input). In fact, our approach can also be incorporated into visual RL locomotion settings, further enhancing their multitask locomotion capabilities.  

\begin{figure*}[htbp]
    \centering
    \vspace{2mm}
    \includegraphics[width=1.0\textwidth]{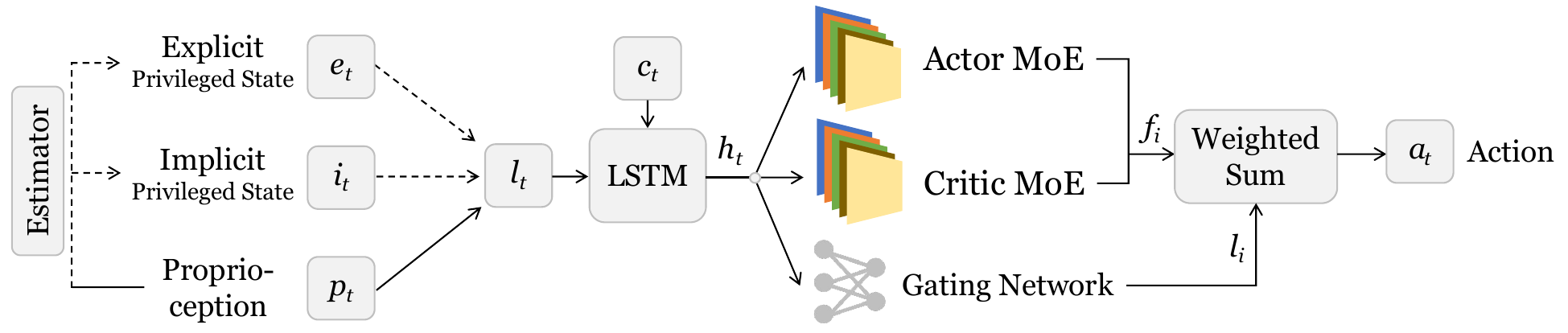}
    \caption{\textbf{Overview of our \texttt{MoELoco} pipeline.} With the design of MoE architecture, our policy achieves robust multitask locomotion ability on various challenging terrains with multiple gaits.}
    \label{fig:pipeline}
    \vspace{-7mm}
\end{figure*}

\subsection{MoE Based Multitask Locomotion Learning}
\vspace{-1.5mm}
\label{sec:moe_learning}
In this section, we introduce our MoE based multitask locomotion learning. We incorporate a two-stage training framework following~\cite{zhu2024robust}, and use PPO~\cite{schulman2017proximal} as our reinforcement learning algorithm. 

\textbf{State Space:} The entire process includes four types of observations: proprioception $\boldsymbol{p}_t$, explicit privileged state $\boldsymbol{e}_t$, implicit privileged state $\boldsymbol{i}_t$, and command $\boldsymbol{c}_t$.  
\textbf{1) Proprioception $\boldsymbol{p}_t$} includes projected gravity and base angular velocity from the IMU, joint positions, joint velocities, and the last action.
\textbf{2) Explicit privileged state $\boldsymbol{e}_t$} contains the base linear velocity (IMU data is too noisy to use) and ground friction.  
\textbf{3) Implicit privileged state $\boldsymbol{i}_t$} includes contact force of different robot link, which must be encoded into a low-dimensional latent representation to mitigate the sim-to-real gap~\cite{zhu2024robust}. 
\textbf{4) Command $\boldsymbol{c}_t$} consists of a velocity command $V = (v_{x}, v_{y}, v_{\text{yaw}})$ and a one-hot vector $g$, where $g = 0$ represents a quadrupedal gait and $g = 1$ represents a bipedal gait in the context of multitask reinforcement learning.

\textbf{Action Space:} The action space $\boldsymbol{a}_t \in \mathbb{R}^{12}$ consists of the desired joint positions for all 12 joints.  

\textbf{Reward Design:} Under the setting of the multitask learning, the robot receives different rewards based on the gait command $g$. For quadrupedal locomotion ($g = 0$), the total reward is defined as $r^{\text{quad}} = r^{\text{quad}}_{\text{track}} + r^{\text{quad}}_{\text{reg}}$. For bipedal locomotion ($g = 1$), the total reward is given by $r^{\text{bip}} = r^{\text{bip}}_{\text{track}} + r^{\text{bip}}_{\text{stand}} + r^{\text{bip}}_{\text{reg}}$. The detailed reward functions can be found in Appendix~\ref{reward_functions}.

\textbf{Termination:} The robot terminates under different circumstances under different gait modes. When $g=0$, the robot terminates when $\theta_{\text{roll}} > 1.0$ or $\theta_{\text{pitch}} > 1.6$. When $g=1$, the robot terminates when any other links except rear feet and calf contacts the ground after 1 second. 

\textbf{Training:} Our training primarily follows the Probability Annealing Selection (PAS) paradigm~\cite{zhu2024saro}. Overall, the robot utilizes both privileged and proprioceptive information in the first stage; but in the second stage, it learns to rely exclusively on proprioception for locomotion, using an estimator to estimate the privileged latent.  In the first training stage, all observation states $[\boldsymbol{p}_t, \boldsymbol{e}_t, \boldsymbol{i}_t, \boldsymbol{c}_t]$ are accessible to train an Oracle policy. The implicit state $\boldsymbol{i}_t$ is first encoded through an encoder network into a latent representation, which is then concatenated with the explicit privileged state $\boldsymbol{e}_t$ and proprioception $\boldsymbol{p}_t$ to form the dual-state representation $\boldsymbol{l}_t = [\text{Enc}(\boldsymbol{i}_t), \boldsymbol{e}_t, \boldsymbol{p}_t]$. Then, the downstream LSTM integrates historical information into state $\boldsymbol{h}_t$. As discussed in \autoref{gd_conflict_main}, jointly learning multiple tasks in multitask reinforcement learning (MTRL) often leads to gradient conflicts. To address this issue, we incorporate the Mixture of Experts (MoE) architecture into both the actor and critic networks, effectively mitigating gradient conflicts and improving learning efficiency. Specifically, each MoE module $f$ operates as follows:

\vspace{-3mm}
\begin{equation}
    \hat{\boldsymbol{g}_{i}} = \text{softmax} (g(\boldsymbol{h}_t))[i],
    \vspace{-1mm}
\end{equation}
\vspace{-4mm}
\begin{equation}
    \boldsymbol{a}_t = \sum_{i = 1}^{N}\hat{\boldsymbol{g}_{i}}\cdot f_{i}(\boldsymbol{h}_t),
    \vspace{-1mm}
\end{equation}

Here, $\boldsymbol{h}_t$ is the output of the low-level LSTM module, $g$ is the gating network that outputs the gating scores, and $f_{i}$ denotes expert $i$. Additionally, we pretrain the estimator module in this stage using a L2 loss $L_{\text{recon}}$ to reconstruct $\mathrm{Estimator}(\boldsymbol{p}_t,\boldsymbol{c}_t)$ into $[\text{Enc}(\boldsymbol{i}_t), \boldsymbol{e}_t]$. In summary, the overall optimization objective is:  
\begin{equation}
    L_{\text{surro}} + L_{\text{value}} + L_{\text{recon}},
    \vspace{-1mm}
\end{equation}  
where $L_{\text{surro}}$ and $L_{\text{value}}$ are surrogate loss and value loss in PPO algorithm.

In the second training stage, the policy can only access $[\boldsymbol{p}_t, \boldsymbol{c}_t]$ as observations. The weights of the estimator, the low-level LSTM, and the MoE modules are initialized by copying them from the first training stage. Probability Annealing Selection~\cite{zhu2024saro} is then employed to gradually adapt the policy to inaccurate estimates with minimal degradation of the Oracle policy performance. Detailed pseudocode is in \autoref{pseudocode}

The MoE architecture facilitates the coordination of similar task skills while minimizing conflicts between heterogeneous tasks by dynamically routing tasks to appropriate experts. This automatic routing enables specialization, improving both efficiency and task performance. Additionally, we incorporate MoE into the critic network to better capture diverse task reward structures. The actor MoE and critic MoE shares the same gating network. By using a shared gating network, we ensure consistency between policy evaluation and action generation.

\subsection{Skill Decomposition and Composition}
A key challenge in multitask learning is how to efficiently utilize previous acquired locomotion skills to form new locomotion tasks. The MoE framework offers a natural solution by dynamically decompose tasks into expert of different skills. Specifically, we conduct quantitative analyses on expert coordination across various tasks. Detailed experiments and results are presented in \autoref{expert_specialization_exp}.

With the automatic decomposition of expert skills, we can recombine them with adjustable weights to synthesize new skills and gaits. Formally, we leverage pretrained experts and modify the gating weights as follows:
\begin{equation}
    \hat{\boldsymbol{g}_{i}} = w[i] \cdot \text{softmax} (g(\boldsymbol{h}_t))[i],
    \vspace{-1mm}
\end{equation}  
where $w[i]$ can be manually defined or dynamically adjusted by a neural network. This formulation enables controlled skill blending, allowing the robot to adapt and generalize to novel locomotion patterns.

Each expert naturally specializes in different aspects of movement, such as balancing, crawling, or obstacle crossing. By selectively adjusting the contributions of these experts, we can construct new locomotion strategies without requiring additional training. This recomposition process highlights the interpretability of MoE-based policies, as each expert’s role can be explicitly identified and manipulated. Detailed experiments are provided in \autoref{skill_composition_exp} and \autoref{additional_exp}.

\section{EXPERIMENTS}
\subsection{Experiment Setup}
We conduct our simulation training in IsaacGym~\cite{makoviychuk2021isaac}, utilizing 4096 robots concurrently on an NVIDIA RTX 3090 GPU. Training begins with 40,000 iterations for plane walking in both gaits, followed by 80,000 iterations on challenging terrain tasks. Finally, we apply PAS for 10,000 iterations to adapt the policy to pure proprioception input. The control frequency in both the simulation environment and the real world is 50Hz. Our policy is deployed on the Unitree Go2 quadruped robot, with an NVIDIA Jetson Orin serving as the onboard computing device. We use PD control for low-level joint execution $(K_p=40.0, K_d=0.5)$. We select expert number $N_{\text{exp}}$ as 6. In all experiment results, \textbf{q} represents \textbf{quadrupedal gait}, and \textbf{b} represents \textbf{bipedal gait}.

\subsection{Multitask Performance}

\begin{table*}[htbp]
    \vspace{2mm}
    \centering
    \caption{\textbf{Quantitative Comparison in Simulation.} Metrics include success rate, average travel distance, and average passing time.}
    \vspace{-1mm}
    \label{tab:combined_comparison}
    \setlength\tabcolsep{8pt}
    \fontsize{8}{10}\selectfont
    \begin{tabular}{c|ccccccccc}
        \toprule
        \multirow{2}{*}{\textbf{Method}} & \multicolumn{9}{c}{\textbf{Success Rate} $\uparrow$} \\
        & \textbf{Mix} & \textbf{Bar (q)} & \textbf{Baffle (q)} & \textbf{Stair (q)} & \textbf{Pit (q)} & \textbf{Slope (q)} & \textbf{Walk (b)} & \textbf{Slope (b)} & \textbf{Stair (b)} \\
        \midrule
        \textbf{Ours} & \textbf{0.879} & \textbf{0.886} & \textbf{0.924} & \textbf{0.684} & \textbf{0.902} & 0.956 & \textbf{0.932} & \textbf{0.961} & \textbf{0.964} \\
        Ours w/o MoE & 0.571 & 0.848 & 0.264 & 0.568 & 0.698 & \textbf{0.988} & 0.826 & 0.504 & 0.453 \\
        RMA & 0.000 & 0.871 & 0.058 & 0.017 & 0.017 & 0.437 & 0.000 & 0.000 & 0.000 \\
        \midrule
        & \multicolumn{9}{c}{\textbf{Average Pass Time (s)} $\downarrow$} \\
        \midrule
        \textbf{Ours} & \textbf{230.98} & \textbf{102.42} & \textbf{87.84} & \textbf{179.14} & \textbf{91.86} & 76.75 & \textbf{92.37} & \textbf{86.14} & \textbf{86.44} \\
        Ours w/o MoE & 315.47 & 125.46 & 318.68 & 214.52 & 161.38 & \textbf{65.28} & 156.76 & 236.67 & 253.62 \\
        RMA & 400.00 & 107.84 & 385.25 & 395.34 & 394.49 & 272.06 & 400.00 & 400.00 & 400.00 \\
        \midrule
        & \multicolumn{9}{c}{\textbf{Average Travel Distance (m)} $\uparrow$} \\
        \midrule
        \textbf{Ours} & \textbf{89.41} & \textbf{28.05} & \textbf{28.02} & 20.42 & \textbf{27.82} & 27.62 & \textbf{27.20} & \textbf{27.99} & \textbf{28.04} \\
        Ours w/o MoE & 57.12 & 27.59 & 17.41 & \textbf{22.66} & 25.59 & \textbf{28.49} & 22.73 & 26.21 & 14.23 \\
        RMA & 13.40 & 27.39 & 11.31 & 3.92 & 12.48 & 21.33 & 2.00 & 2.00 & 2.00 \\
        \bottomrule
    \end{tabular}
    \vspace{-7mm}
\end{table*}

\subsubsection{Simulation Experiment}
We conduct comparative simulation experiments with the following baselines: 
\begin{itemize}
    \item \textbf{Ours w/o MoE}~\cite{zhu2024robust}: Uses the same framework as ours but replaces the MoE module with a simple MLP. We control the total parameter to be the same as our MoE policy.
    \item \textbf{RMA}~\cite{kumar2021rma}: Employs a 1D-CNN as an asynchronous adaptation module within a teacher-student training framework, without using an MoE module.  
\end{itemize}  

We constructed a benchmark for quadrupedal robot locomotion across different tasks. Our benchmark consists of a $5m \times 100m$ runway with various obstacles evenly distributed along the path. The obstacles include:

\begin{table}[h]
    \centering
    \vspace{-2mm}
    \caption{Benchmark tasks for simulation experiments}
    \label{tab:benchmark}
    \begin{tabular}{c|c|c}
        \toprule
        \textbf{Obstacle Type} & \textbf{Specification} & \textbf{Gait Mode} \\
        \midrule
        Bars       & 5 bars, height: 0.05m -- 0.2m  & Quadrupedal \\
        Pits       & 5 pits, width: 0.05m -- 0.2m   & Quadrupedal \\
        Baffles    & 5 baffles, height: 0.3m -- 0.22m & Quadrupedal \\
        Up Stairs  & 3 sets, step height: 5cm -- 15cm  & Quadrupedal \\
        Down Stairs & 3 sets, step height: 5cm -- 15cm & Quadrupedal \\
        Up Slopes  & 3 sets, incline: $10^{\circ}$ -- $35^{\circ}$ & Quadrupedal \\
        Down Slopes & 3 sets, incline: $10^{\circ}$ -- $35^{\circ}$ & Quadrupedal \\
        Plane & 10m flat surface & Bipedal \\
        Up Slopes  & 3 sets, incline: $10^{\circ}$ -- $35^{\circ}$ & Bipedal \\
        Down Slopes & 3 sets, incline: $10^{\circ}$ -- $35^{\circ}$ & Bipedal \\
        Down Stairs & 3 sets, step height: 5cm -- 15cm & Bipedal \\
        \bottomrule
    \end{tabular}
    \vspace{-3mm}
\end{table}

We also conducted experiments for each challenging tasks, each track is 30 meters long. We consider three metrics: \textbf{Success Rate}, \textbf{Average Pass Time}, and \textbf{Average Travel Distance}. A trial is considered successful if the robot reaches within 1m of the target point within 400 seconds. Upon completion, we record the travel time. Failure cases include falling off the runway, getting stuck, or meeting the termination conditions outlined in \autoref{sec:moe_learning}. For failed trials, the pass time is recorded as 400 seconds. We compute the overall success rate and average pass time across all trials. Additionally, we measure the average lateral travel distance of all robots at the end of the evaluation.

As shown in \autoref{tab:combined_comparison}, our MoE policy achieves the best performance in the mixed-task benchmark across all three metrics. In all single-task evaluations, except for quadrupedal slope walking, our policy outperforms others. This exception may be attributed to the relative simplicity of quadrupedal slope walking. Furthermore, policy without MoE struggles to effectively traverse the challenging multitask terrain setting, as it is significantly affected by gradient conflicts, as discussed in \autoref{gd_conflict_main}. Regarding RMA, we adhere to its original implementation, which utilizes only an MLP backbone and a CNN encoder. This design choice leads to its suboptimal performance on multiple challenging terrains.

\subsubsection{Real World Experiments}
We deploy our MoE policy zero-shotly on real robots and conduct real world experiments. We test mix terrain that contains all challenging tasks, as well as each separated terrains. For the mix terrain, our robot first need to consequently cross 20cm bar, 22cm baffle, 15cm stairs, 20cm pits and 30 degree slopes in a quadrupedal gait. Then it receives a bipedal command to stand up, walk up the 30 degree slope, turn around and walk down. For each tests, we test for 20 trails and record the average success rate. We also test different policies for single tasks. 

\begin{figure}[h]
    \centering
    \vspace{-4mm}
    \includegraphics[width=\linewidth]{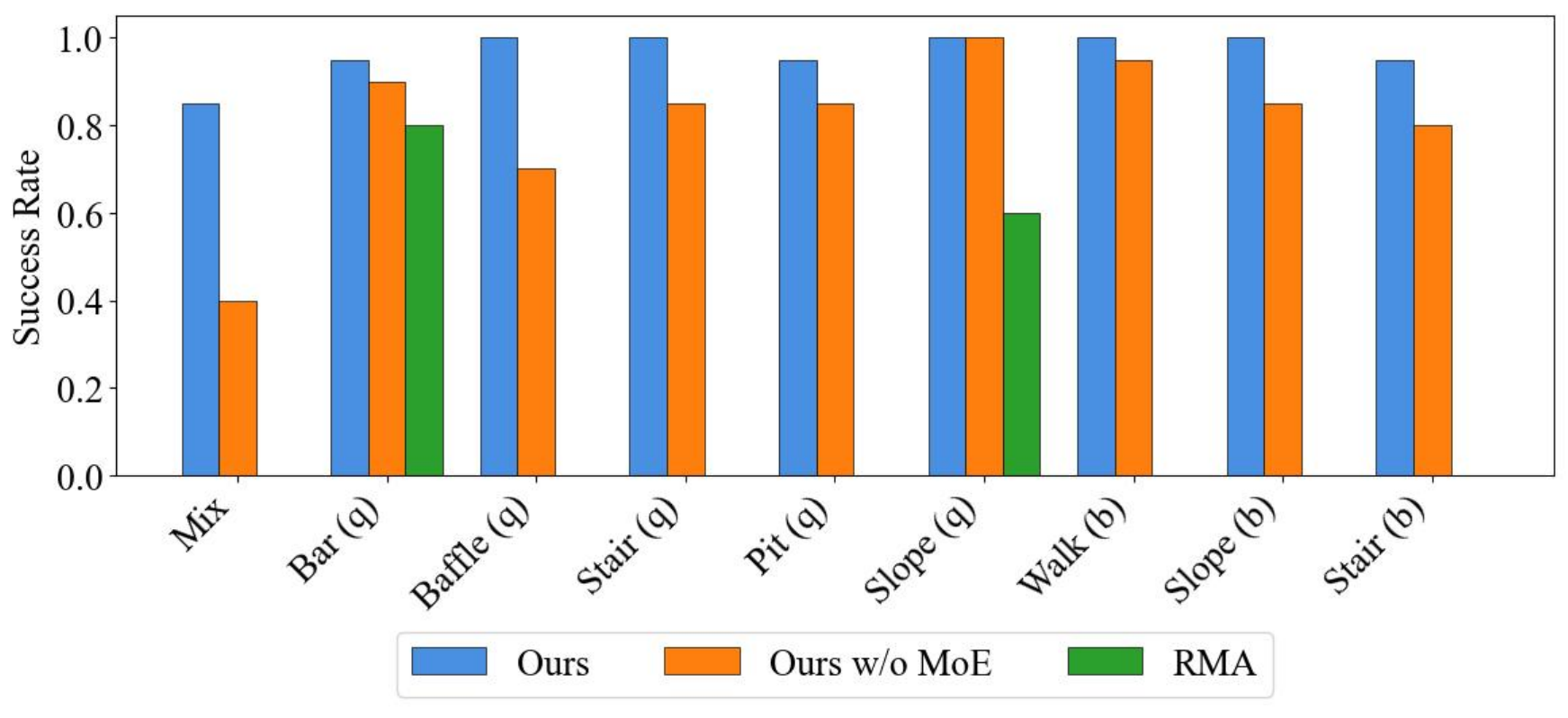}
    \vspace{-6.5mm}
    \caption{\textbf{Real world success rate over multiple terrains and gaits.}}
    \label{fig:real-sr}
    \vspace{-4mm}
\end{figure}

\begin{figure}[h]
    \centering
    \vspace{-3mm}
    \includegraphics[width=\linewidth]{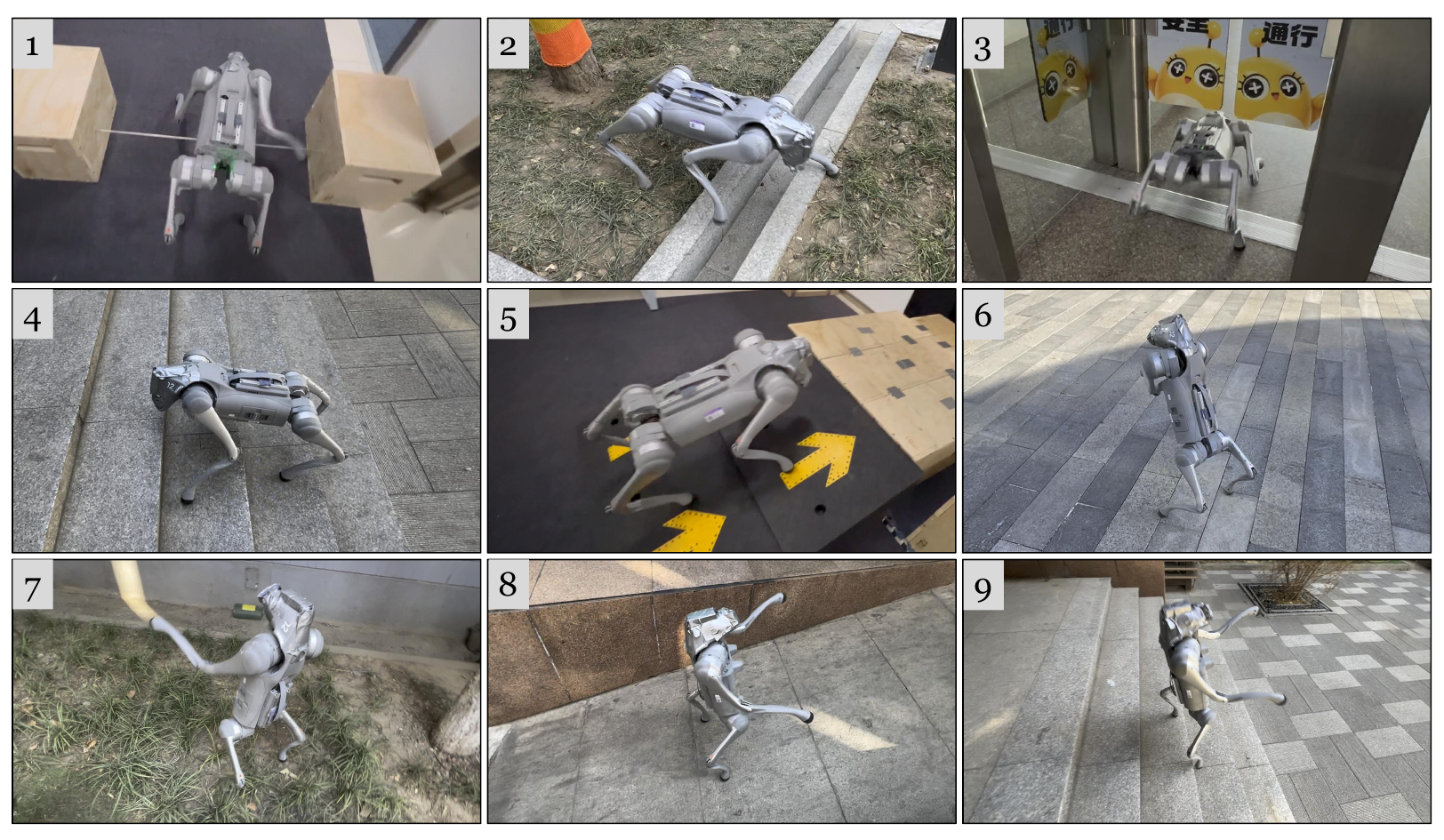}
    \vspace{-5.5mm}
    \caption{\textbf{Real-world experiments over multiple terrains and gaits:} 1. Bar (Quad), 2. Pit (Quad), 3. Baffle (Quad), 4. Stair (Quad), 5. Slope (Quad), 6. Stand up (Bip), 7. Walk (Bip), 8. Slope (Bip), 9. Stair (Bip).}
    \label{fig:real-exp}
    \vspace{-5mm}
\end{figure}

As shown in \autoref{fig:real-sr}, our method achieves better performance across all types of tasks and demonstrates a significantly higher success rate in mix terrain. Additionally, we conduct experiments in more outdoor environments, as shown in \autoref{fig:real-exp}, further demonstrating its robustness and generalization.

\subsection{Gradient Conflict Alleviation}\label{gd_conflict_main}
In order to unveil whether gradient conflict can be reduced by applying mixture of experts, we conducted gradient conflict experiments. We resume from the checkpoints after pretraining for 15000 epochs and run the training process of multitask for 500 epochs of 4096 quadrupedal robots. We average the gradient throughout the process. We consider two metrics to measure the gradient conflict of robot locomotion. \textbf{1) Cosine Similarity:} We compute the normalized dot product of all parameters' gradients of different tasks. Smaller cosine similarity indicates larger gradient conflicts. \textbf{2) Negative gradient ratio:} We compute negative gradient update ratio for each pair of task gradients. Larger negative gradient ratio indicates larger gradient conflict. We test the gradient conflict between 5 different tasks: quadrupedal bar crossing, quadrupedal baffle crawling, quadrupedal stair walking, bipedal plane walking and bipedal slope walking. 

\begin{table}[htbp]
    \vspace{-2mm}
    \centering
    \caption{\textbf{Cosine similarity of gradients of different tasks.} Left represents MoE policy and right represents standard policy.}
    \vspace{-1mm}
    \label{tab:cos_dist}
    \setlength\tabcolsep{2pt}
    \fontsize{7}{9}\selectfont
    \begin{tabular}{c|ccccccc}
        \toprule
        \multirow{2}{*}{\textbf{MoE/Standard}}
        & \multicolumn{5}{c}{\textbf{Gradient Cosine Similarity } $\uparrow$} \\
        & \textbf{Bar (q)} &  \textbf{Baffle (q)} & \textbf{Stair (q)} &  \textbf{Slope Up (b)} & \textbf{Slope down (b)} \\
         \midrule
         \textbf{Bar (q)} & - & \textbf{0.519}/0.474 & \textbf{0.606}/0.592 & \textbf{0.278}/-0.132 & \textbf{0.091}/-0.128 \\
        \textbf{Baffle (q)} & - & - & 0.369/\textbf{0.384} & \textbf{0.062}/-0.091 & \textbf{0.061}/-0.101 \\
        \textbf{Stair (q)} & - & - & - & \textbf{0.046}/-0.023 & \textbf{0.052}/0.015 \\
        \textbf{Slope up (b)} & - & - & - & - & \textbf{0.806}/0.709 \\
        \textbf{Slope down (b)} & - & - & - & - & - \\
        \bottomrule
    \end{tabular}
    \vspace{-1.5mm}
\end{table}

\begin{table}[htbp]
    \vspace{-4mm}
    \centering
    \caption{\textbf{Negative entry ratio of MoE and Standard policy on different tasks.} Left represents MoE policy and right represents standard policy.}
    \vspace{-1mm}
    \label{tab:neg_entry}
    \setlength\tabcolsep{2pt}
    \fontsize{7}{9}\selectfont
    \begin{tabular}{c|ccccccc}
        \toprule
        \multirow{2}{*}{\textbf{MoE/Standard}}
        & \multicolumn{5}{c}{\textbf{Gradient Negative Entries (\%)} $\downarrow$} \\
        & \textbf{Bar (q)} &  \textbf{Baffle (q)} & \textbf{Stair (q)} &  \textbf{Slope Up (b)} & \textbf{Slope down (b)} \\
         \midrule
         \textbf{Bar (q)} & - & \textbf{35.72}/37.33 & 32.67/\textbf{32.62} & \textbf{45.50}/ 55.12 & \textbf{49.83}/50.80 \\
        \textbf{Baffle (q)} & - & - & 39.90/\textbf{38.52} & \textbf{49.86}/55.91 & \textbf{49.91}/51.68 \\
        \textbf{Stair (q)} & - & - & - & \textbf{49.52}/50.15 & \textbf{50.04}/50.34 \\
        \textbf{Slope up (b)} & - & - & - & - & \textbf{23.17}/30.91 \\
        \textbf{Slope down (b)} & - & - & - & - & - \\
        \bottomrule
    \end{tabular}
    \vspace{-4mm}
\end{table}

As shown in \autoref{tab:cos_dist} and \autoref{tab:neg_entry}, the MoE policy significantly reduces gradient conflict between bipedal and quadrupedal tasks. It also minimizes gradient conflict even between quadrupedal tasks that require fundamentally different skills, such as quadrupedal bar crossing and quadrupedal baffle crawling.

\subsection{Training Performance}
\begin{figure}[h]
    \centering
    \vspace{-4mm}
    \includegraphics[width=\linewidth]{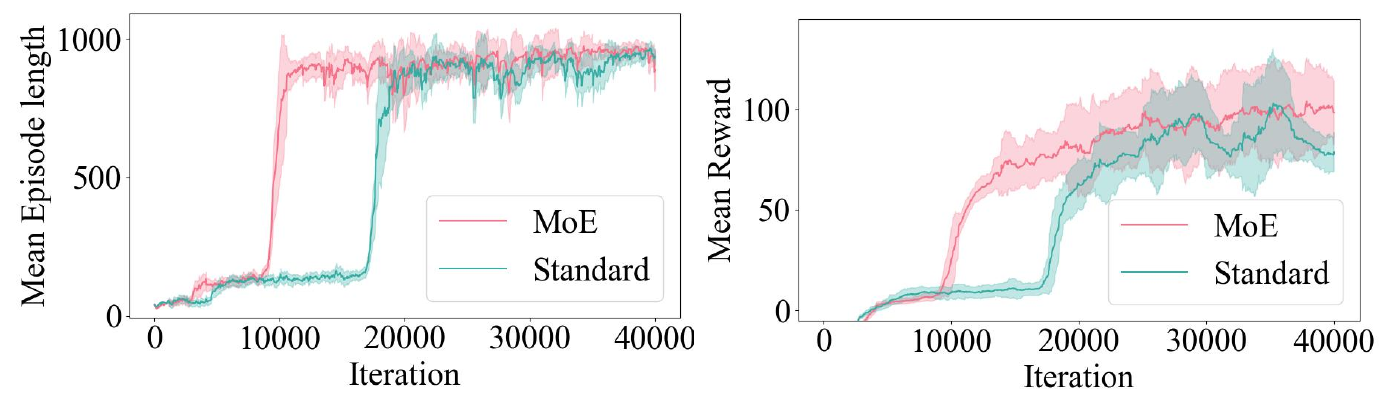}
    \vspace{-5.5mm}
    \caption{\textbf{Training curve of our multitask policy in the pretraining stage.}}
    \label{fig:training_curve}
    \vspace{-4mm}
\end{figure}
In terms of training performance, we focus on the mean reward and mean episode length. The mean reward reflects the policy's ability to exploit the environment, while the mean episode length indicates how well the robot learns to stand and walk. We present the training curve during the plane pretraining stage, where the robot learns both bipedal and quadrupedal plane walking. As shown in \autoref{fig:training_curve}, our MoE policy outperforms the standard policy with similar total parameters across both metrics.

\subsection{Expert Specialization Analysis}\label{expert_specialization_exp}

\begin{figure*}[h]
    \centering
    \includegraphics[width=\linewidth]{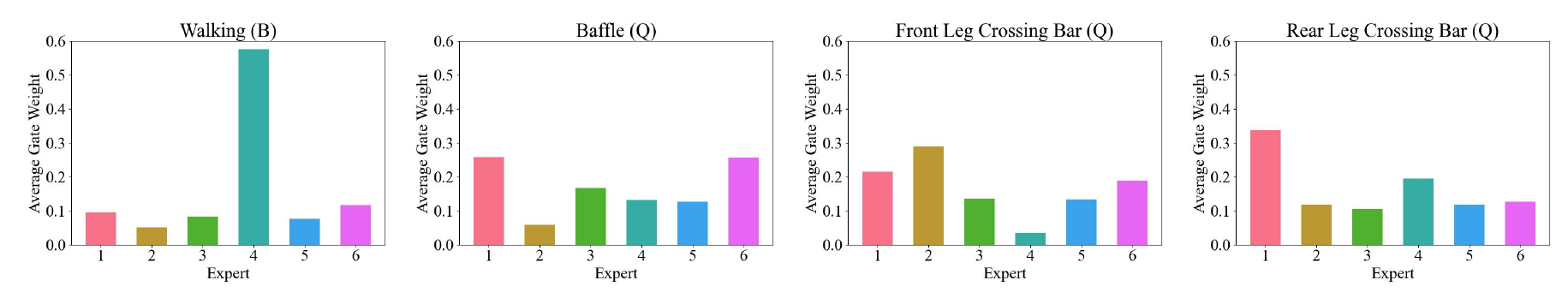}
    \vspace{-6.5mm}
    \caption{\textbf{Expert usage in different tasks.} From left to right is: Bipedal walking, Quadrupedal Baffle crawling, Front Leg Crossing Bars and Rear Leg Crossing Bars.}
    \label{fig:gating_weight_plot}
    \vspace{-7mm}
\end{figure*}

\begin{figure}[h]
    \centering
    \includegraphics[width=0.9\linewidth]{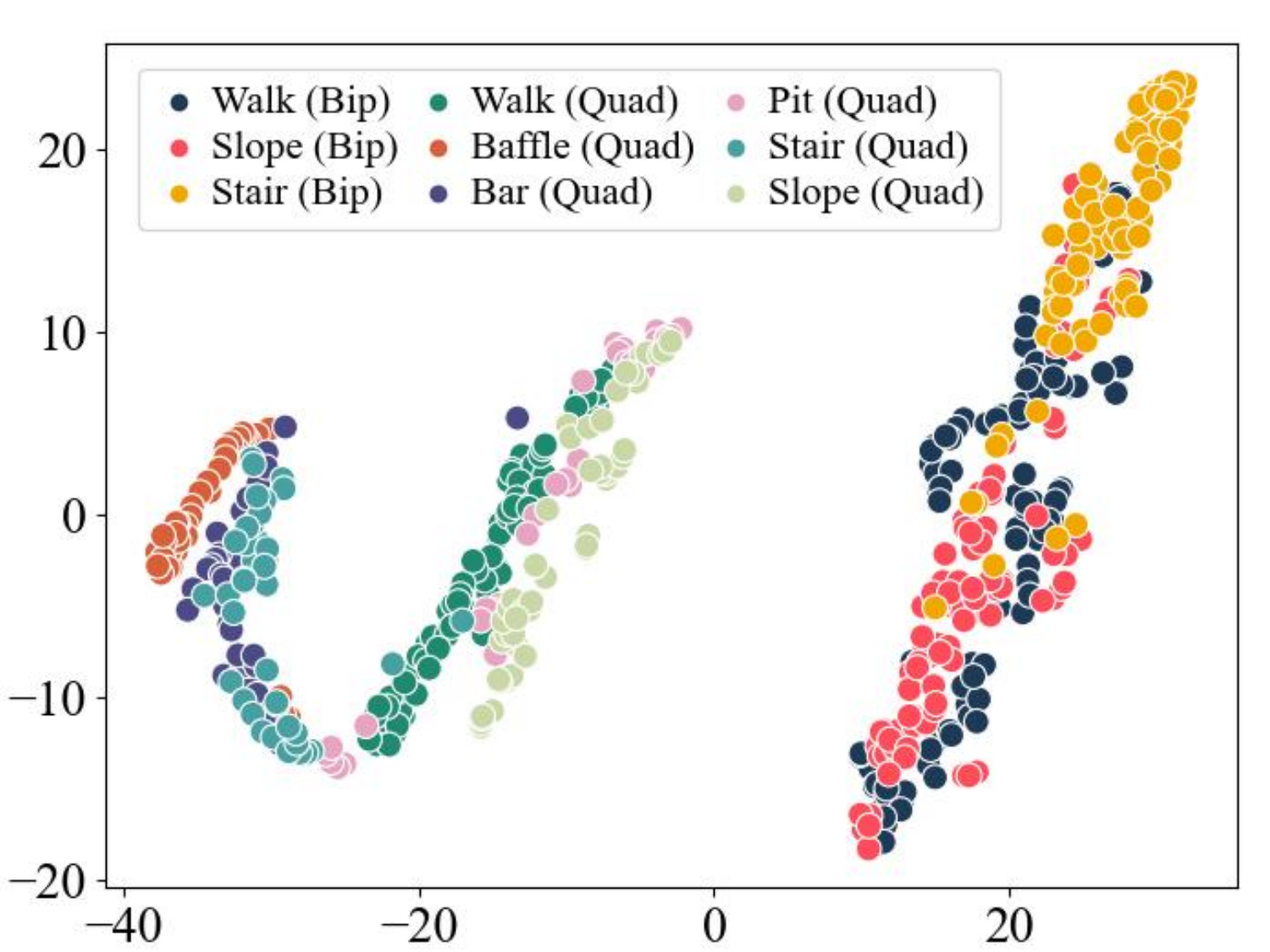}
    \vspace{-1.5mm}
    \caption{\textbf{t-SNE result of gating network output on different terrains and gaits.}}
    \label{fig:expert_gating_tsne}
    \vspace{-7mm}
\end{figure}

We conduct both qualitative and quantitative experiments to analyze the emergent composition of skills. As shown in \autoref{fig:gating_weight_plot}, we plot the mean weight of different experts across various tasks. It is evident that the distribution of gating weights varies significantly from task to task, demonstrating the expertise and differentiation of various experts.

To further analyze the composition of different experts across various tasks, we use t-SNE to visualize the output of the gating network for different tasks (i.e., the weight of each expert). As shown in \autoref{fig:expert_gating_tsne}, the bipedal and quadrupedal tasks form distinct clusters. Quadrupedal slope walking and quadrupedal pit crossing tasks are performed using a gait similar to quadrupedal plane walking, and thus, they cluster closely together. In contrast, bar crossing, baffle crawling, and stair climbing exhibit a gait that differs more significantly, resulting in them clustering further apart.

\vspace{-1mm}
\subsection{Skill Composition}\label{skill_composition_exp}

\begin{figure}[h]
    \centering
    \vspace{-5mm}
    \includegraphics[width=\linewidth]{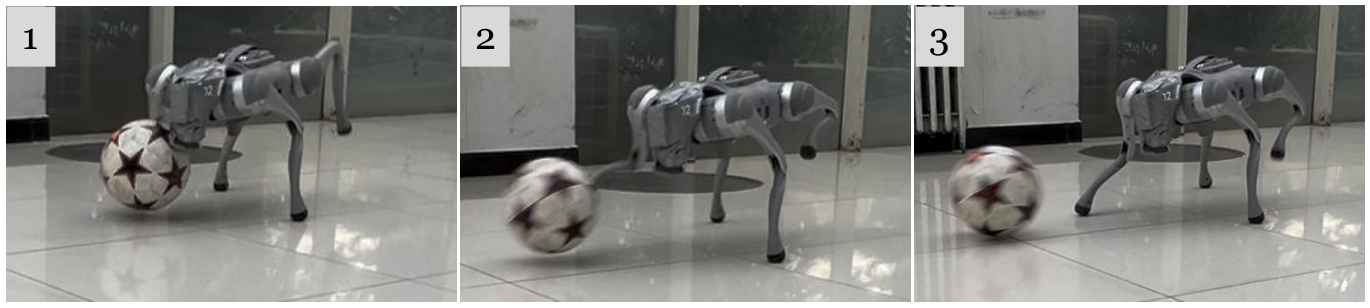}
    \vspace{-5.5mm}
    \caption{\textbf{Manually designed new dribbling gait by selecting two experts.}}
    \label{fig:dribble}
    \vspace{-4mm}
\end{figure}

During our training, we found that our experts not only specialize and cooperate across different tasks, but they also emerge with specific, human-interpretable skills. We discovered that one expert specializes in balancing, which helps lift the robot’s body but limits its agility. Another expert is responsible for lifting one of the front legs to perform crossing tasks, enabling the robot to execute basic movement skills. By selecting the balancing expert and the crossing expert, we are able to zero-shot transfer to a new dribbling pattern. In this gait, we select the two experts mentioned above, manually double the gating weight of the crossing expert, and mask out all other experts. As shown in \autoref{fig:dribble}, the new dribbling gait allows the robot to walk effectively while periodically using one of its front legs to kick the ball. This skill composition results from the automatic skill decomposition and interpretability inherent in the MoE architecture. In contrast, a standard neural network would function as a black box, lacking such interpretable skill decomposition.

\vspace{-0.5mm}
\subsection{Additional Experiment}\label{additional_exp}

\begin{figure}[h]
    \centering
    \vspace{-4.5mm}
    \includegraphics[width=\linewidth]{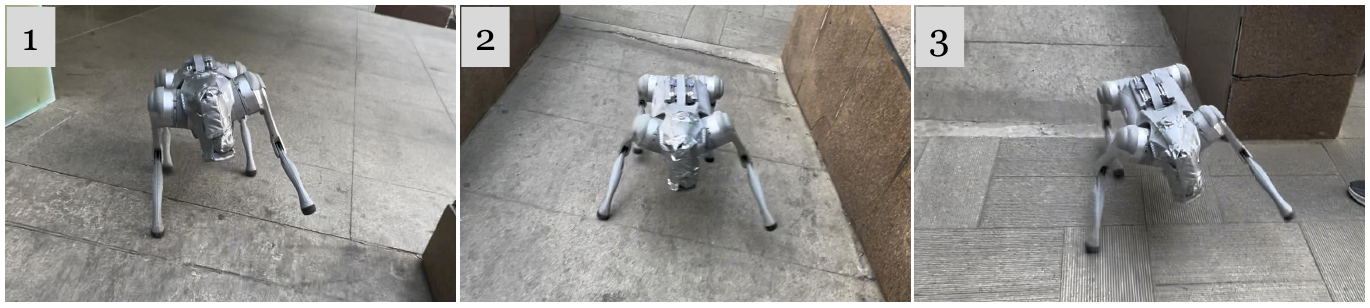}
    \vspace{-6mm}
    \caption{\textbf{\texttt{MoE-Loco} can quickly adapt to a three-footed gait by training a new expert.} 1) ground plane, 2) slope up, and 3) slope down.}
    \label{fig:3feet}
    \vspace{-2mm}
\end{figure}

We conduct an adaptation learning experiment to demonstrate how our pretrained experts can be recomposed and adapted to new tasks. In this experiment, we design the robot to walk on three feet. We introduce a newly initialized expert, freeze the parameters of the original experts, and update only the gating network. As shown in \autoref{fig:3feet}, the robot can walk on both flat ground and a slope using only three feet. The newly added expert only needs to learn how to lift one leg, while leveraging the walking and slope capabilities of the original experts.

\section{CONCLUSION}

We introduced MoE-Loco, a multitask locomotion framework that utilizes a Mixture of Experts architecture to train a single policy for quadrupedal robots. Our approach effectively mitigates gradient conflicts, leading to improved training efficiency and task performance. Through extensive evaluations in both simulated and real-world environments, we demonstrated the capability of our method to handle a variety of terrains and gaits. Future work will explore extending this approach to incorporate sensory perception such as camera and Lidar to enhance adaptability in more complex tasks.





\section*{APPENDIX}

\subsection{Reward Functions}\label{reward_functions}
\vspace{-1.5mm}
The robots in bipedal gait receives different reward to the robots in quadrupedal gait. Detailed explanation is shown in \autoref{tab:reward}.

\begin{table}[htbp]
    \renewcommand{\arraystretch}{1.0}
    \centering
    \vspace{-4mm}
    \caption{Reward functions}
    \label{tab:reward}
    \setlength\tabcolsep{2pt}
    \fontsize{4.7}{6.7}\selectfont
    \begin{tabular}{c|c|c|c}
        \toprule
         \textbf{Type} & \textbf{Item} & \textbf{Formula} & \textbf{Weight}\\ \midrule
        \multirow{4}{*}{\textbf{Quadrupedal Tracking}} 
         & Tracking lin vel & $\exp\left(-\frac{\left\|\mathbf{v}_{c,xy} - \mathbf{v}_{b,xy}\right\|^2}{\mathbf{\sigma}}\right)$ & $7.0$ \\
         & Tracking ang vel & $\exp\left(-\frac{\left(\mathbf{c}_{\text{yaw}} - \mathbf{\omega}_{\text{yaw}}\right)^2}{\mathbf{\sigma}}\right)$ & $2.5$ \\
         & Termination & $-1$ & $-1.0$ \\
         & Alive & $1$ & $1.0$ \\ \midrule
        \multirow{11}{*}{\textbf{Quadrupedal Regularization}} 
         & Joint pos & $\left( \mathbf{q} - \mathbf{q}_{\text{default}} \right)^2$ & $-0.05$ \\
         & Joint vel & $\left\|\dot{\mathbf{q}}\right\|_2$ & $-0.002$ \\
         & Joint acc & $\left\|\ddot{\mathbf{q}}\right\|_2$ & $-2\times10^{-6}$ \\
         & Ang vel stability & $\left(\left\|\mathbf{\omega}_{t,x}\right\|_2 + \left\|\mathbf{\omega}_{t,y}\right\|_2\right)$ & $-0.2$ \\
         & Feet in air & $\prod_{i=1}^{\mathbf{N}}\mathbb{I}\Bigl\{\mathbf{F}^{\text{foot}}_{i,z} < 1\Bigr\} \;+\; \sum_{i=1}^{\mathbf{N}}\mathbb{I}\Bigl\{\mathbf{F}^{\text{foot}}_{i,z} < 1\Bigr\}\,\mathbb{I}\Bigl\{\mathbf{F}^{\text{calf}}_{i,z} \ge 1\Bigr\}$ & $-0.05$ \\ 
         & Front hip pos & $\sum_{i\in \mathcal{I}_{\text{front hip}}}\Bigl(\mathbf{q}_i - \mathbf{q}_i^{\text{default}}\Bigr)^2$ & $-0.2$ \\ 
         & Rear hip pos & $\sum_{i\in \mathcal{I}_{\text{rear hip}}}\Bigl(\mathbf{q}_i - \mathbf{q}_i^{\text{default}}\Bigr)^2$ & $-0.5$ \\ 
         & Base height & $\Bigl(\mathbf{z}_{\text{base}} - \frac{1}{\mathbf{N}}\sum_{i=1}^{\mathbf{N}} \mathbf{h}_i - \mathbf{h}_{\text{target}}\Bigr)^2$ & $-0.1$ \\
         & Balance & $\left\|\mathbf{F}_{\text{feet},0} + \mathbf{F}_{\text{feet},2} - \mathbf{F}_{\text{feet},1} - \mathbf{F}_{\text{feet},3}\right\|_2$ & $-2\times10^{-5}$ \\
         & Joint limit & $\left\|\Bigl( \mathbf{q} < \mathbf{q}_{\min} \Bigr) \lor \Bigl( \mathbf{q} > \mathbf{q}_{\max} \Bigr)\right\|_{1}$ & $-0.01$ \\
         & Torque exceed limit & $\sum_{i=1}^{\mathbf{N}_{\text{sub}}} \sum_{j=1}^{12} \max\Bigl\{ \,|\mathbf{\tau}_{ij}| - \mathbf{\tau}^{\text{limit}}_j,\; 0 \Bigr\}$ & $-2.0$ \\ \midrule
\multirow{2}{*}{\textbf{Bipedal Stand}} 
        & Orientation & $\Bigl(0.5\,\cos\theta + 0.5\Bigr)^2,\quad \theta = \arccos\Bigl(\frac{\mathbf{g}\cdot\mathbf{t}}{\|\mathbf{g}\|\|\mathbf{t}\|}\Bigr)$ & $1.0$ \\
        & Base height linear & $\min\Bigl(\max\Bigl(\frac{\mathbf{z}_{\text{root}} - T_{\min}}{T_{\max}-T_{\min}},\,0\Bigr),\,1\Bigr)$ & $0.8$ \\ \midrule
\multirow{4}{*}{\textbf{Bipedal Tracking}} 
         & Tracking lin vel & $\exp\Bigl(-\frac{\left\|\mathbf{v}_{c} - \mathbf{v}_{a}\right\|^2}{\mathbf{\sigma}}\Bigr)\,\mathbb{I}\Bigl\{\cos\mathbf{\theta} > 0.95\Bigr\}\,\frac{\mathbf{z}_{\mathrm{base}} - \mathbf{s}_{\mathrm{low}}}{\mathbf{s}_{\mathrm{high}} - \mathbf{s}_{\mathrm{low}}}$ & $3.0$ \\
         & Tracking ang vel & $\exp\Bigl(-\frac{\left(\mathbf{c}_{\mathrm{roll}} - \mathbf{\omega}_{\mathrm{roll}}\right)^2}{\mathbf{\sigma}_{ang}}\Bigr)\,\mathbb{I}\Bigl\{\cos\mathbf{\theta} > 0.95\Bigr\}\,\frac{\mathbf{z}_{\mathrm{base}} - \mathbf{s}_{\mathrm{low}}}{\mathbf{s}_{\mathrm{high}} - \mathbf{s}_{\mathrm{low}}}$ & $2.5$ \\
         & Termination & $-1$ & $-1.0$ \\
         & Alive & $1$ & $1.0$ \\\midrule
\multirow{14}{*}{\textbf{Bipedal Regularization}} 
         & Rear air & $\prod_{i\in \mathbf{R}}\mathbb{I}\Bigl\{\mathbf{f}^{\mathrm{rear}}_{i,z} < 1\Bigr\} \;+\; \sum_{i\in \mathbf{R}}\mathbb{I}\Bigl\{\mathbf{f}^{\mathrm{rear}}_{i,z} < 1\Bigr\}\,\mathbb{I}\Bigl\{\mathbf{f}^{\mathrm{rear-calf}}_{i,z} \ge 1\Bigr\}$ & $-0.5$ \\
         & Front hip pos & $\sum_{i\in \mathbf{F}_H}\Bigl(\mathbf{q}_i - \mathbf{q}_i^{\mathrm{default}}\Bigr)^2$ & $-0.1$ \\
         & Rear hip pos & $\sum_{i\in \mathbf{R}_H}\Bigl(\mathbf{q}_i - \mathbf{q}_i^{\mathrm{default}}\Bigr)^2$ & $-0.18$ \\
         & Rear pos balance & $\left\|\mathbf{q}^{\mathrm{rear\,left}}_{i} - \mathbf{q}^{\mathrm{rear\,right}}_{i}\right\|_{2}$ & $-0.05$ \\
         & Front joint pos & $\mathbf{1}\Bigl\{\mathbf{t} > \mathbf{T}_{\mathrm{allow}}\Bigr\}\sum_{i\in \mathbf{F}}\Bigl(\mathbf{q}_i - \mathbf{q}_i^{\mathrm{default}}\Bigr)^2$ & $-0.2$ \\
         & Front joint vel & $\mathbf{1}\Bigl\{\mathbf{t} > \mathbf{T}_{\mathrm{allow}}\Bigr\}\sum_{i\in \mathbf{F}}\dot{\mathbf{q}}_i^2$ & $-1\times10^{-3}$ \\
         & Front joint acc & $\mathbf{1}\Bigl\{\mathbf{t} > \mathbf{T}_{\mathrm{allow}}\Bigr\}\sum_{i\in \mathbf{F}}\Bigl(\frac{\Delta\dot{\mathbf{q}}_i}{\Delta \mathbf{t}}\Bigr)^2$ & $-2\times10^{-6}$ \\
         & Legs energy substeps & $\frac{1}{\mathbf{N}_s}\sum_{j=1}^{\mathbf{N}_s}\sum_{i=1}^{\mathbf{N}_{\mathrm{dof}}} \Bigl(\mathbf{\tau}_{ij}\,\dot{\mathbf{q}}_{ij}\Bigr)^2$ & $-1\times10^{-6}$ \\
         & Torque exceed limits & $\sum_{j}\sum_{i}\max\Bigl\{|\mathbf{\tau}_{ij}| - \mathbf{\tau}^{\mathrm{limit}}_i,\; 0\Bigr\}$ & $-2.0$ \\
         & Joint limits & $\frac{1}{\mathbf{N}_s}\sum_{j}\sum_{i}\mathbb{I}\Bigl\{\mathbf{q}_{ij}\notin [\mathbf{q}_i^{\min},\mathbf{q}_i^{\max}]\Bigr\}$ & $-0.06$ \\
         & Collision & $\sum_{k}\mathbb{I}\Bigl\{\|\mathbf{f}_k\| > 1\Bigr\}$ & $-2.0$ \\
         & Action rate & $\left\|\mathbf{a}^{\mathrm{last}}_i - \mathbf{a}_i\right\|_2$ & $-0.03$ \\
         & Joint vel & $\left\|\dot{\mathbf{q}}\right\|_2$ & $-2\times10^{-3}$ \\
         & Joint acc & $\left\|\ddot{\mathbf{q}}\right\|_2$ & $-3\times10^{-6}$ \\
    \bottomrule
    \end{tabular}
    \vspace{-9.0mm}
\end{table}

\subsection{Algorithm Pseudocode}\label{pseudocode}
\vspace{-6.5mm}
\begin{algorithm}[H]
\fontsize{7.5}{9.5}\selectfont
\caption{Training Stage 1}\label{alg:stage_1}
\begin{algorithmic}[1]
    \For{total iteration}
        \State Initialize rollout buffer $\mathcal{D} \gets \emptyset$
        \For{num steps}
            \State $\mathbf{z}_t \gets \mathrm{Enc}(\mathbf{i}_t)$
            \State $\hat{\mathbf{l}}_t \gets [\mathrm{Estimator}(\mathbf{p}_t,\mathbf{c}_t),\mathbf{p}_t]$
            \State $\mathbf{l}_t \gets [\mathbf{z}_t,\mathbf{e}_t,\mathbf{p}_t]$
            \State $\mathbf{h}_t \gets \mathrm{LSTM}([\mathbf{l}_t,\mathbf{c}_t])$
            \State $\hat{\mathbf{g}} \gets \mathrm{softmax}(g(\mathbf{h}_t))$
            \State $\mathbf{a}_t \gets \sum_{i=1}^{N}\hat{\mathbf{g}}_i \cdot f_i(\mathbf{h}_t)$
            \State Execute $\mathbf{a}_t$, observe reward $r_t$ and next state
            \State Reset if terminated
            \State $t \gets t+1$
            \State Store rollout in $\mathcal{D}$
        \EndFor
        \State Compute PPO losses: $L_{\mathrm{surro}},\,L_{\mathrm{value}}$
        \State Compute reconstruction loss:
        \State \quad $L_{\mathrm{recon}} = \sum_{\substack{\hat{\mathbf{l}}_{i},\,\mathbf{l}_{i} \in \mathcal{D}}} \left\|\hat{\mathbf{l}}_{i} - \mathbf{l}_{i}\right\|^{2}$
        \State $L = L_{\mathrm{surro}} + L_{\mathrm{value}} + L_{recon}$
        \State Update policy and value network
    \EndFor
    \State \Return Oracle Policy
\end{algorithmic}
\end{algorithm}

\vspace*{-6.0mm}
\begin{algorithm}[H]
\fontsize{7.5}{9.5}\selectfont
\caption{Training Stage 2}\label{alg:stage_2}
\begin{algorithmic}[1]
    \State Copy parameters from Oracle Policy
    \For{total iteration}
        \State Initialize rollout buffer $\mathcal{D} \gets \emptyset$
        \For{num steps}
            \State $\hat{\mathbf{l}}_t \gets [\mathrm{Estimator}(\mathbf{p}_t,\mathbf{c}_t),\,\mathbf{p}_t]$
            \State $\mathbf{l}_t \gets [\mathbf{z}_t,\mathbf{e}_t,\mathbf{p}_t]$
            \State $\mathbf{P}_t \gets \alpha^{\text{t}}$
            \State $\bar{\mathbf{l}}_t \gets \mathrm{Probability\ Selection}(\mathbf{P}_t,\hat{\mathbf{l}}_t,\mathbf{l}_t)$
            \State $\mathbf{h}_t \gets \mathrm{LSTM}([\bar{\mathbf{l}}_t,\,\mathbf{c}_t])$
            \State $\hat{\mathbf{g}} \gets \mathrm{softmax}(g(\mathbf{h}_t))$
            \State $\mathbf{a}_t \gets \sum_{i=1}^{N}\hat{\mathbf{g}}_i \cdot f_i(\mathbf{h}_t)$
            \State Execute $\mathbf{a}_t$, observe reward $r_t$ and next state
            \State Reset if terminated
            \State $t \gets t+1$
            \State Store rollout in $\mathcal{D}$
        \EndFor
        \State Compute PPO losses: $L_{\mathrm{surro}},\,L_{\mathrm{value}}$
        \State Compute reconstruction loss:
        \State \quad $L_{\mathrm{recon}} = \sum_{\substack{\hat{\mathbf{l}}_{i},\,\mathbf{l}_{i} \in \mathcal{D}}} \left\|\hat{\mathbf{l}}_{i} - \mathbf{l}_{i}\right\|^{2}$
        \State $L = L_{\mathrm{surro}} + L_{\mathrm{value}} + L_{recon}$
        \State Update policy and value network
    \EndFor
    \State \Return Final Policy
\end{algorithmic}
\end{algorithm}
\vspace{-4mm}
\subsection{Network Architecture Details}\label{nn_arch_detail}
The architecture of different modules used in our experiment is shown in \autoref{tab:network}.
\begin{table}[htbp]
    \vspace{-4mm}
    \renewcommand{\arraystretch}{1.0}
    \fontsize{6}{8}\selectfont
    \caption{\label{tab:network}Network architecture details}
    \vspace{-1mm}
    \centering
    \begin{tabular}{c|cc}
        \toprule
        \textbf{Network} & \textbf{Type}  & \textbf{Dims} \\ \midrule
        Actor RNN & LSTM & [256] \\
        Critic RNN & LSTM & [256] \\
        Estimator Module & LSTM & [256] \\
        Estimator Latent Encoder & MLP & [256, 128]\\
        Implicit Encoder & MLP & [32, 16] \\
        Expert Head & MLP & [256, 128, 128]\\
        Standard Head & MLP & [640, 640, 128]\\
        Gating Network & MLP & [128] \\
        \bottomrule
    \end{tabular}
    \vspace{-6mm}
\end{table}

\subsection{Domain Randomization}\label{domain_rand}

We introduce dynamic randomization to ensure our policy can be safely transfered to real environment. The random ranges are shown in the \autoref{tab:random}.

\begin{table}[htbp]
    \vspace{1mm}
    \renewcommand{\arraystretch}{1.0}
    \fontsize{6}{8}\selectfont
    \centering
    \caption{\label{tab:random}Domain randomization}
    \vspace{-1mm}
    \begin{tabular}{c|cc}
        \toprule
        \textbf{Parameters} & \textbf{Range} & \textbf{Unit} \\ \midrule
        Base mass & [1, 3] & $kg$\\
        Mass position of X axis & [-0.2, 0.2] & $m$\\
        Mass position of Y axis & [-0.1, 0.1] & $m$\\
        Mass position of Z axis & [-0.05, 0.05] & $m$\\
        Friction & [0, 2] & - \\
        Initial joint positions & [0.5, 1.5]$~\times~$nominal value & $rad$\\
        Motor strength & [0.9, 1.1]$~\times~$nominal value & - \\
        Proprioception latency & [0.005, 0.045] & $s$\\
        \bottomrule
    \end{tabular}
    \vspace{-1mm}
\end{table}

Gaussian noise are added to the input observation to simulate the noise in the real world, as shown in ~\autoref{tab:gaussian}. 

\begin{table}[htbp]
    \vspace{-3mm}
    \renewcommand{\arraystretch}{1.0}
    \fontsize{6}{8}\selectfont
    \centering
    \caption{\label{tab:gaussian}Gaussian noise}
    \vspace{-1mm}
    \begin{tabular}{c|cc}
        \toprule
        \textbf{Observation} & \textbf{Gaussian Noise Amplitude} & \textbf{Unit} \\ \midrule
        Linear velocity & 0.05 & $m/s$\\
        Angular velocity & 0.2 & $rad/s$\\
        Gravity & 0.05 & $m/s^2$\\
        Joint position & 0.01 & $rad$\\
        Joint velocity & 1.5 & $rad/s$\\
        \bottomrule
    \end{tabular}
    \vspace{-7mm}
\end{table}
\subsection{Training Details}\label{training_details}
In order to prevent legs dragging on the floor, and achieve a robust performance on uneven terrains, we apply fractal noise~\cite{zhuang2023robot} to the ground. We use maximum noise scale $\text{z}_{max} = 0.1$. We use PPO as our reinforcement learning algorithm. The hyperparameters are shown in \autoref{tab:ppohyper}.

\begin{table}[h!]
    \vspace{-4mm}
    \renewcommand{\arraystretch}{1.0}
    \setlength\tabcolsep{18pt}
    \fontsize{6}{8}\selectfont
    \caption{PPO hyperparameters}
    \vspace{-1.5mm}
    \centering
    \label{tab:ppohyper}
    \begin{tabular}{c|c}
        \toprule
        \textbf{Hyperparameter} & \textbf{Value}  \\ \midrule
        clip min std & 0.05 \\
        clip param & 0.2\\
        gamma & 0.99\\
        lam & 0.95\\
        desired kl & 0.01\\
        entropy coef & 0.01\\
        learning rate & 0.001\\
        max grad norm & 1\\
        num mini batch & 4\\
        num steps per env & 24\\
        \bottomrule
    \end{tabular}
    \vspace{-6.5mm}
\end{table}




\bibliographystyle{IEEEtran}
\bibliography{references}

\begin{thebibliography}{10}
\providecommand{\url}[1]{#1}
\csname url@rmstyle\endcsname
\providecommand{\newblock}{\relax}
\providecommand{\bibinfo}[2]{#2}
\providecommand\BIBentrySTDinterwordspacing{\spaceskip=0pt\relax}
\providecommand\BIBentryALTinterwordstretchfactor{4}
\providecommand\BIBentryALTinterwordspacing{\spaceskip=\fontdimen2\font plus
\BIBentryALTinterwordstretchfactor\fontdimen3\font minus \fontdimen4\font\relax}
\providecommand\BIBforeignlanguage[2]{{%
\expandafter\ifx\csname l@#1\endcsname\relax
\typeout{** WARNING: IEEEtran.bst: No hyphenation pattern has been}%
\typeout{** loaded for the language `#1'. Using the pattern for}%
\typeout{** the default language instead.}%
\else
\language=\csname l@#1\endcsname
\fi
#2}}

\bibitem{kumar2023cascaded}
K.~N. Kumar, I.~Essa, and S.~Ha, ``Cascaded compositional residual learning for complex interactive behaviors,'' \emph{IEEE Robotics and Automation Letters}, vol.~8, no.~8, pp. 4601--4608, 2023.

\bibitem{klipfel2023learning}
A.~Klipfel, N.~Sontakke, R.~Liu, and S.~Ha, ``Learning a single policy for diverse behaviors on a quadrupedal robot using scalable motion imitation,'' in \emph{2023 IEEE/RSJ International Conference on Intelligent Robots and Systems (IROS)}.\hskip 1em plus 0.5em minus 0.4em\relax IEEE, 2023, pp. 2768--2775.

\bibitem{kumar2021rma}
A.~Kumar, Z.~Fu, D.~Pathak, and J.~Malik, ``Rma: Rapid motor adaptation for legged robots,'' \emph{arXiv preprint arXiv:2107.04034}, 2021.

\bibitem{su2024leveraging}
Z.~Su, X.~Huang, D.~Ordo{\~n}ez-Apraez, Y.~Li, Z.~Li, Q.~Liao, G.~Turrisi, M.~Pontil, C.~Semini, Y.~Wu, \emph{et~al.}, ``Leveraging symmetry in rl-based legged locomotion control,'' \emph{arXiv preprint arXiv:2403.17320}, 2024.

\bibitem{ji2022concurrent}
G.~Ji, J.~Mun, H.~Kim, and J.~Hwangbo, ``Concurrent training of a control policy and a state estimator for dynamic and robust legged locomotion,'' \emph{IEEE Robotics and Automation Letters}, vol.~7, no.~2, pp. 4630--4637, 2022.

\bibitem{long2024learning}
J.~Long, J.~Ren, M.~Shi, Z.~Wang, T.~Huang, P.~Luo, and J.~Pang, ``Learning humanoid locomotion with perceptive internal model,'' \emph{arXiv preprint arXiv:2411.14386}, 2024.

\bibitem{zhuang2023robot}
Z.~Zhuang, Z.~Fu, J.~Wang, C.~Atkeson, S.~Schwertfeger, C.~Finn, and H.~Zhao, ``Robot parkour learning,'' \emph{arXiv preprint arXiv:2309.05665}, 2023.

\bibitem{wu2023learning}
J.~Wu, G.~Xin, C.~Qi, and Y.~Xue, ``Learning robust and agile legged locomotion using adversarial motion priors,'' \emph{IEEE Robotics and Automation Letters}, 2023.

\bibitem{luo2024pie}
S.~Luo, S.~Li, R.~Yu, Z.~Wang, J.~Wu, and Q.~Zhu, ``Pie: Parkour with implicit-explicit learning framework for legged robots,'' \emph{arXiv preprint arXiv:2408.13740}, 2024.

\bibitem{cheng2024extreme}
X.~Cheng, K.~Shi, A.~Agarwal, and D.~Pathak, ``Extreme parkour with legged robots,'' in \emph{2024 IEEE International Conference on Robotics and Automation (ICRA)}.\hskip 1em plus 0.5em minus 0.4em\relax IEEE, 2024, pp. 11\,443--11\,450.

\bibitem{liu2023improving}
S.~Liu, Z.~Chen, Y.~Liu, Y.~Wang, D.~Yang, Z.~Zhao, Z.~Zhou, X.~Yi, W.~Li, W.~Zhang, \emph{et~al.}, ``Improving generalization in visual reinforcement learning via conflict-aware gradient agreement augmentation,'' in \emph{Proceedings of the IEEE/CVF International Conference on Computer Vision}, 2023, pp. 23\,436--23\,446.

\bibitem{zhou2022convergence}
S.~Zhou, W.~Zhang, J.~Jiang, W.~Zhong, J.~Gu, and W.~Zhu, ``On the convergence of stochastic multi-objective gradient manipulation and beyond,'' \emph{Advances in Neural Information Processing Systems}, vol.~35, pp. 38\,103--38\,115, 2022.

\bibitem{jacobs1991adaptive}
R.~A. Jacobs, M.~I. Jordan, S.~J. Nowlan, and G.~E. Hinton, ``Adaptive mixtures of local experts,'' \emph{Neural computation}, vol.~3, no.~1, pp. 79--87, 1991.

\bibitem{obando2024mixtures}
J.~Obando-Ceron, G.~Sokar, T.~Willi, C.~Lyle, J.~Farebrother, J.~Foerster, G.~K. Dziugaite, D.~Precup, and P.~S. Castro, ``Mixtures of experts unlock parameter scaling for deep rl,'' \emph{arXiv preprint arXiv:2402.08609}, 2024.

\bibitem{li2024mixtures}
K.~Li, M.~Cucuringu, L.~S{\'a}nchez-Betancourt, and T.~Willi, ``Mixtures of experts for scaling up neural networks in order execution,'' in \emph{Proceedings of the 5th ACM International Conference on AI in Finance}, 2024, pp. 669--676.

\bibitem{celik2024acquiring}
O.~Celik, A.~Taranovic, and G.~Neumann, ``Acquiring diverse skills using curriculum reinforcement learning with mixture of experts,'' \emph{arXiv preprint arXiv:2403.06966}, 2024.

\bibitem{makoviychuk2021isaac}
V.~Makoviychuk, L.~Wawrzyniak, Y.~Guo, M.~Lu, K.~Storey, M.~Macklin, D.~Hoeller, N.~Rudin, A.~Allshire, A.~Handa, \emph{et~al.}, ``Isaac gym: High performance gpu-based physics simulation for robot learning,'' \emph{arXiv preprint arXiv:2108.10470}, 2021.

\bibitem{mittal2023orbit}
M.~Mittal, C.~Yu, Q.~Yu, J.~Liu, N.~Rudin, D.~Hoeller, J.~L. Yuan, R.~Singh, Y.~Guo, H.~Mazhar, A.~Mandlekar, B.~Babich, G.~State, M.~Hutter, and A.~Garg, ``Orbit: A unified simulation framework for interactive robot learning environments,'' \emph{IEEE Robotics and Automation Letters}, vol.~8, no.~6, pp. 3740--3747, 2023.

\bibitem{hwangbo2019learning}
J.~Hwangbo, J.~Lee, A.~Dosovitskiy, D.~Bellicoso, V.~Tsounis, V.~Koltun, and M.~Hutter, ``Learning agile and dynamic motor skills for legged robots,'' \emph{Science Robotics}, vol.~4, no.~26, p. eaau5872, 2019.

\bibitem{zhu2024robust}
S.~Zhu, R.~Huang, L.~Mou, and H.~Zhao, ``Robust robot walker: Learning agile locomotion over tiny traps,'' \emph{arXiv preprint arXiv:2409.07409}, 2024.

\bibitem{lee2020learning}
J.~Lee, J.~Hwangbo, L.~Wellhausen, V.~Koltun, and M.~Hutter, ``Learning quadrupedal locomotion over challenging terrain,'' \emph{Science robotics}, vol.~5, no.~47, p. eabc5986, 2020.

\bibitem{margolis2024rapid}
G.~B. Margolis, G.~Yang, K.~Paigwar, T.~Chen, and P.~Agrawal, ``Rapid locomotion via reinforcement learning,'' \emph{The International Journal of Robotics Research}, vol.~43, no.~4, pp. 572--587, 2024.

\bibitem{he2024agile}
T.~He, C.~Zhang, W.~Xiao, G.~He, C.~Liu, and G.~Shi, ``Agile but safe: Learning collision-free high-speed legged locomotion,'' \emph{arXiv preprint arXiv:2401.17583}, 2024.

\bibitem{li2024learning}
Y.~Li, J.~Li, W.~Fu, and Y.~Wu, ``Learning agile bipedal motions on a quadrupedal robot,'' in \emph{2024 IEEE International Conference on Robotics and Automation (ICRA)}.\hskip 1em plus 0.5em minus 0.4em\relax IEEE, 2024, pp. 9735--9742.

\bibitem{smith2023learning}
L.~Smith, J.~C. Kew, T.~Li, L.~Luu, X.~B. Peng, S.~Ha, J.~Tan, and S.~Levine, ``Learning and adapting agile locomotion skills by transferring experience,'' \emph{arXiv preprint arXiv:2304.09834}, 2023.

\bibitem{cheng2024quadruped}
Y.~Cheng, H.~Liu, G.~Pan, L.~Ye, H.~Liu, and B.~Liang, ``Quadruped robot traversing 3d complex environments with limited perception,'' \emph{arXiv preprint arXiv:2404.18225}, 2024.

\bibitem{hoeller2024anymal}
D.~Hoeller, N.~Rudin, D.~Sako, and M.~Hutter, ``Anymal parkour: Learning agile navigation for quadrupedal robots,'' \emph{Science Robotics}, vol.~9, no.~88, p. eadi7566, 2024.

\bibitem{caruana1997multitask}
R.~Caruana, ``Multitask learning,'' \emph{Machine learning}, vol.~28, pp. 41--75, 1997.

\bibitem{collobert2008unified}
R.~Collobert and J.~Weston, ``A unified architecture for natural language processing: Deep neural networks with multitask learning,'' in \emph{Proceedings of the 25th international conference on Machine learning}, 2008, pp. 160--167.

\bibitem{vandenhende2021multi}
S.~Vandenhende, S.~Georgoulis, W.~Van~Gansbeke, M.~Proesmans, D.~Dai, and L.~Van~Gool, ``Multi-task learning for dense prediction tasks: A survey,'' \emph{IEEE transactions on pattern analysis and machine intelligence}, vol.~44, no.~7, pp. 3614--3633, 2021.

\bibitem{liu2016recurrent}
P.~Liu, X.~Qiu, and X.~Huang, ``Recurrent neural network for text classification with multi-task learning,'' \emph{arXiv preprint arXiv:1605.05101}, 2016.

\bibitem{pinto2017learning}
L.~Pinto and A.~Gupta, ``Learning to push by grasping: Using multiple tasks for effective learning,'' in \emph{2017 IEEE international conference on robotics and automation (ICRA)}.\hskip 1em plus 0.5em minus 0.4em\relax IEEE, 2017, pp. 2161--2168.

\bibitem{yu2020gradient}
T.~Yu, S.~Kumar, A.~Gupta, S.~Levine, K.~Hausman, and C.~Finn, ``Gradient surgery for multi-task learning,'' \emph{Advances in Neural Information Processing Systems}, vol.~33, pp. 5824--5836, 2020.

\bibitem{liu2021conflict}
B.~Liu, X.~Liu, X.~Jin, P.~Stone, and Q.~Liu, ``Conflict-averse gradient descent for multi-task learning,'' \emph{Advances in Neural Information Processing Systems}, vol.~34, pp. 18\,878--18\,890, 2021.

\bibitem{huang2024mentor}
S.~Huang, Z.~Zhang, T.~Liang, Y.~Xu, Z.~Kou, C.~Lu, G.~Xu, Z.~Xue, and H.~Xu, ``Mentor: Mixture-of-experts network with task-oriented perturbation for visual reinforcement learning,'' \emph{arXiv preprint arXiv:2410.14972}, 2024.

\bibitem{finn2017model}
C.~Finn, P.~Abbeel, and S.~Levine, ``Model-agnostic meta-learning for fast adaptation of deep networks,'' in \emph{International conference on machine learning}.\hskip 1em plus 0.5em minus 0.4em\relax PMLR, 2017, pp. 1126--1135.

\bibitem{duan2016rl}
Y.~Duan, J.~Schulman, X.~Chen, P.~L. Bartlett, I.~Sutskever, and P.~Abbeel, ``Rl\textsuperscript{2}: Fast reinforcement learning via slow reinforcement learning,'' \emph{arXiv preprint arXiv:1611.02779}, 2016.

\bibitem{sodhani2021multi}
S.~Sodhani, A.~Zhang, and J.~Pineau, ``Multi-task reinforcement learning with context-based representations,'' in \emph{International Conference on Machine Learning}.\hskip 1em plus 0.5em minus 0.4em\relax PMLR, 2021, pp. 9767--9779.

\bibitem{yang2020multi}
C.~Yang, K.~Yuan, Q.~Zhu, W.~Yu, and Z.~Li, ``Multi-expert learning of adaptive legged locomotion,'' \emph{Science Robotics}, vol.~5, no.~49, p. eabb2174, 2020.

\bibitem{ze2023gnfactor}
Y.~Ze, G.~Yan, Y.-H. Wu, A.~Macaluso, Y.~Ge, J.~Ye, N.~Hansen, L.~E. Li, and X.~Wang, ``Gnfactor: Multi-task real robot learning with generalizable neural feature fields,'' in \emph{Conference on Robot Learning}.\hskip 1em plus 0.5em minus 0.4em\relax PMLR, 2023, pp. 284--301.

\bibitem{shafiee2024manyquadrupeds}
M.~Shafiee, G.~Bellegarda, and A.~Ijspeert, ``Manyquadrupeds: Learning a single locomotion policy for diverse quadruped robots,'' in \emph{2024 IEEE International Conference on Robotics and Automation (ICRA)}.\hskip 1em plus 0.5em minus 0.4em\relax IEEE, 2024, pp. 3471--3477.

\bibitem{yang2020multiloco}
C.~Yang, K.~Yuan, Q.~Zhu, W.~Yu, and Z.~Li, ``Multi-expert learning of adaptive legged locomotion,'' \emph{Science Robotics}, vol.~5, no.~49, p. eabb2174, 2020.

\bibitem{shah2023mtac}
N.~Shah, K.~Tiwari, and A.~Bera, ``Mtac: Hierarchical reinforcement learning-based multi-gait terrain-adaptive quadruped controller,'' \emph{arXiv preprint arXiv:2401.03337}, 2023.

\bibitem{jordan1994hierarchical}
M.~I. Jordan and R.~A. Jacobs, ``Hierarchical mixtures of experts and the em algorithm,'' \emph{Neural computation}, vol.~6, no.~2, pp. 181--214, 1994.

\bibitem{deisenroth2015distributed}
M.~Deisenroth and J.~W. Ng, ``Distributed gaussian processes,'' in \emph{International conference on machine learning}.\hskip 1em plus 0.5em minus 0.4em\relax PMLR, 2015, pp. 1481--1490.

\bibitem{lepikhin2020gshard}
D.~Lepikhin, H.~Lee, Y.~Xu, D.~Chen, O.~Firat, Y.~Huang, M.~Krikun, N.~Shazeer, and Z.~Chen, ``Gshard: Scaling giant models with conditional computation and automatic sharding,'' \emph{arXiv preprint arXiv:2006.16668}, 2020.

\bibitem{jiang2024mixtral}
A.~Q. Jiang, A.~Sablayrolles, A.~Roux, A.~Mensch, B.~Savary, C.~Bamford, D.~S. Chaplot, D.~d.~l. Casas, E.~B. Hanna, F.~Bressand, \emph{et~al.}, ``Mixtral of experts,'' \emph{arXiv preprint arXiv:2401.04088}, 2024.

\bibitem{liu2024deepseek}
A.~Liu, B.~Feng, B.~Xue, B.~Wang, B.~Wu, C.~Lu, C.~Zhao, C.~Deng, C.~Zhang, C.~Ruan, \emph{et~al.}, ``Deepseek-v3 technical report,'' \emph{arXiv preprint arXiv:2412.19437}, 2024.

\bibitem{zhang2024m3oe}
Z.~Zhang, S.~Liu, J.~Yu, Q.~Cai, X.~Zhao, C.~Zhang, Z.~Liu, Q.~Liu, H.~Zhao, L.~Hu, \emph{et~al.}, ``M3oe: Multi-domain multi-task mixture-of experts recommendation framework,'' in \emph{Proceedings of the 47th International ACM SIGIR Conference on Research and Development in Information Retrieval}, 2024, pp. 893--902.

\bibitem{jiang2023adamct}
J.~Jiang, P.~Zhang, Y.~Luo, C.~Li, J.~B. Kim, K.~Zhang, S.~Wang, X.~Xie, and S.~Kim, ``Adamct: adaptive mixture of cnn-transformer for sequential recommendation,'' in \emph{Proceedings of the 32nd ACM international conference on information and knowledge management}, 2023, pp. 976--986.

\bibitem{gou2023mixture}
Y.~Gou, Z.~Liu, K.~Chen, L.~Hong, H.~Xu, A.~Li, D.-Y. Yeung, J.~T. Kwok, and Y.~Zhang, ``Mixture of cluster-conditional lora experts for vision-language instruction tuning,'' \emph{arXiv preprint arXiv:2312.12379}, 2023.

\bibitem{chen2024llava}
S.~Chen, Z.~Jie, and L.~Ma, ``Llava-mole: Sparse mixture of lora experts for mitigating data conflicts in instruction finetuning mllms,'' \emph{arXiv preprint arXiv:2401.16160}, 2024.

\bibitem{schulman2017proximal}
J.~Schulman, F.~Wolski, P.~Dhariwal, A.~Radford, and O.~Klimov, ``Proximal policy optimization algorithms,'' \emph{arXiv preprint arXiv:1707.06347}, 2017.

\bibitem{zhu2024saro}
S.~Zhu, D.~Li, L.~Mou, Y.~Liu, N.~Xu, and H.~Zhao, ``Saro: Space-aware robot system for terrain crossing via vision-language model,'' \emph{arXiv preprint arXiv:2407.16412}, 2024.

\end{thebibliography}

\balance

\end{document}